\documentclass[conference,a4paper]{IEEEtran}
\IEEEoverridecommandlockouts
\usepackage{cite}
\usepackage{amsmath,amssymb,amsfonts}
\usepackage{algorithmic}
\usepackage{graphicx}
\usepackage{textcomp}
\usepackage{xcolor}
\def\BibTeX{{\rm B\kern-.05em{\sc i\kern-.025em b}\kern-.08em
    T\kern-.1667em\lower.7ex\hbox{E}\kern-.125emX}}
\begin{document}

\title{Sensor Fusion by Spatial Encoding\\
for Autonomous Driving\\
}


\author{
    \IEEEauthorblockN{Quoc-Vinh Lai-Dang$^1$, Jihui Lee$^2$, Bumgeun Park$^1$, Dongsoo Har$^{1, 2}$}
    \IEEEauthorblockA{$^1$\textit{Cho Chun Shik Graduate School of Mobility}}
    \IEEEauthorblockA{$^2$\textit{Division of Future Vehicle}}
    \textit{Korea Advanced Institute of Science and Technology (KAIST)}
    \\Daejeon, South Korea
    \\ldqvinh, jihui, j4t123, dshar@kaist.ac.kr
}

\maketitle

\begin{abstract}
Sensor fusion is critical to perception systems for task domains such as autonomous driving and robotics. Recently, the Transformer integrated with CNN has demonstrated high performance in sensor fusion for various perception tasks. In this work, we introduce a method for fusing data from camera and LiDAR. By employing Transformer modules at multiple resolutions, proposed method effectively combines local and global contextual relationships. The performance of the proposed method is validated by extensive experiments with two adversarial benchmarks with lengthy routes and high-density traffics. The proposed method outperforms previous approaches with the most challenging benchmarks, achieving significantly higher driving and infraction scores. Compared with TransFuser, it achieves 8\% and 19\% improvement in driving scores for the Longest6 and Town05 Long benchmarks, respectively.
\end{abstract}

\begin{IEEEkeywords}
Sensor Fusion, Autonomous Driving, Transformer
\end{IEEEkeywords}

\section{Introduction}
Multi-modal perception methods that fuse data from camera and LiDAR \cite{b1,b2} have made significant advancements in the field of autonomous driving. Although the LiDAR is excellent at understanding the geometric properties of 3D scenes \cite{lai2022learning}, its limited ability to detect semantic objects \cite{b3}, such as traffic lights, makes it difficult to be used in practice. In contrast, camera provides semantic information \cite{rajendran2022relmobnet, rajendran2022lightweight}, but they do not have 3D depth perception. Therefore, integrating camera and LiDAR, while making up for each other's weakness, is crucial in autonomous driving. Unlike channel-dependent data fusion with multiple wireless sensors \cite{Park1} optionally capable of data decoding \cite{Park2}, entire concern with sensor fusion for autonomous driving is limited to establishing fusion mechanism.

Previous methods have successfully combined features from the local neighborhood by incorporating sensor data in a projected space \cite{b4, b17}. Interactions with distant traffic lights and signs pose challenges for methods based on local information. To address this, there is a need to bridge the gap between sensor data processing and the utilization of both semantic and spatial information. Recently, there has been notable advancement in the Transformer architectures \cite{b5, b6}, which has emerged as a viable alternative to CNN not just in wireless sensor networks \cite{kim2020machine}, but also in the area of autonomous driving. In this paper, we propose a novel approach employing attention mechanisms in the Transformer architecture \cite{b7} to aggregate sensor data features. Main contributions of our work are three-fold:

\begin{itemize}
\item Our approach combines sinusoidal positional and learnable sensor encodings, yielding a refined feature representation for multi-modal fusion.
\item The fusion mechanism boosts safety and interpretability in autonomous driving scenarios, contributing to more reliable decision-making.
\item Proposed approach achieves superior performance with two challenging CARLA benchmarks, namely Longest6 and Town05 Long.
\end{itemize}

\begin{figure*}[htbp]
\centerline{\includegraphics[width=17cm]{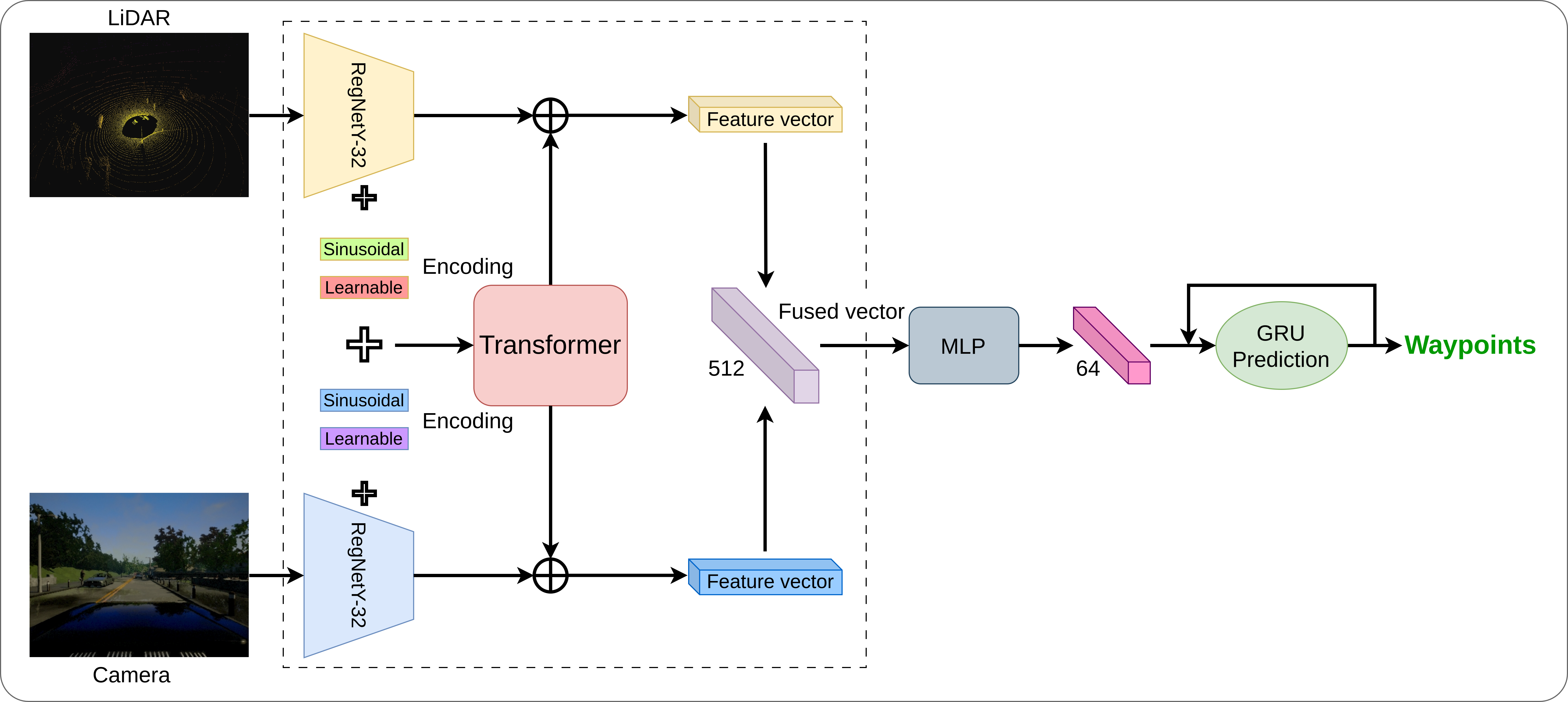}}
\caption{Fusion processes of proposed method. CNNs are used to extract features from multi-modal sensors. The features are fused in a Transformer encoder at multiple resolutions. The resulting 512-dimensional feature vector is a compact representation of the environment. It is processed with an MLP and passed to a waypoint prediction network.}
\label{fig1}
\end{figure*}

\section{Related Works}
Multi-sensor fusion has become increasingly popular in 3D detection. Based on when the different sensors are fused, current methods for multi-sensor fusion can be classified into three categories: detection-level fusion, point-level fusion, and proposal-level fusion.

\subsection{Detection-level fusion}
Detection-level fusion, also known as late-fusion, is a straightforward approach to combining the sensor data from multiple sensors. The model generates Bird's Eye View (BEV) detections for each sensor and aggregates and deduplicates them. However, this approach does not fully utilize the unique characteristics provided by sensors. The Camera-LiDAR Object Candidates fusion (CLOCs) \cite{b9} addresses this limitation by effectively combining the strengths of each modality. It performs 2D and 3D detection using cameras and LiDAR, and removes false positives using geometric consistency.

\subsection{Point-level fusion}
Point-level fusion \cite{b10}, also referred to as early-fusion, combines data from LiDAR point clouds with features extracted from camera images. This involves enhancing LiDAR points with camera pixels using the transformation matrix, but camera-to-LiDAR projections can result in semantic loss due to sparsity, which constrains fusion quality.

\subsection{Proposal-level fusion}
Notable works such as Multi-View 3D networks (MV3D) \cite{b11} propose initial bounding boxes using LiDAR features, and refine them iteratively using camera features. The BEVFusion \cite{b12} generates BEV features from camera images and fuses them with LiDAR features in the BEV space. The TransFuser \cite{b13} uses Transformers to fuse single-view images and LiDAR BEV representations, resulting in a compact representation of local and global context. 

This paper introduces novel techniques to capture local and global relationships with multiple sensors, addressing challenges that previous methods could not solve.
\begin{table*}[htbp]
\caption{Longest6 Benchmark Results}
\begin{center}
\begin{tabular}{c|ccc|cccccccc}
\textbf{Method} & \textbf{DS $\uparrow$} & \textbf{RC $\uparrow$} & \textbf{IS $\uparrow$} & \textbf{Ped $\downarrow$} & \textbf{Veh$\downarrow$ } & \textbf{Stat$\downarrow$} & \textbf{Red $\downarrow$} & \textbf{OR $\downarrow$} & \textbf{Dev $\downarrow$} & \textbf{TO $\downarrow$} & \textbf{Block $\downarrow$} \\ \hline
Latent TransFuser &37.31 &\bfseries95.18  &0.38 &0.03 &1.05 &0.37 &1.28 &0.47 &0.88 &0.08 &0.20 \\
Late Fusion &22.47 &83.30  &0.27 &0.05 &4.63 &0.28 &0.11 &0.48 &0.02 &0.11 &0.21 \\
Geometric Fusion &27.32 &91.13 &0.30 &0.06 &4.64 &0.17 &0.13 &0.48 &\bfseries0.00 &0.05 &\bfseries0.11 \\
TransFuser &41.86 &74.36 &0.63  &0.01 &0.38 &\bfseries0.08 &0.07 &\bfseries0.04 &\bfseries0.00 &\bfseries0.04 &0.48 \\
\hline
Ours &\bfseries45.64 &73.60 &\bfseries0.65 &\bfseries0.00 &\bfseries0.35 &0.14 &\bfseries0.06 &0.10 &\bfseries0.00 &0.07 &0.29 \\ \hline
\end{tabular}
\label{tab:1}
\end{center}
\end{table*}

\begin{table}[htbp]
\caption{Town05 Long Benchmark Results}
\begin{center}
\begin{tabular}{c|ccc}
\textbf{Method} & \textbf{DS $\uparrow$} & \textbf{RC $\uparrow$} & \textbf{IS $\uparrow$} \\ \hline
Latent TransFuser  &49.42 &89.06 &0.59 \\
Late Fusion &57.81 &98.24 &0.58 \\
Geometric Fusion &55.32 &\bfseries98.39 &0.57 \\
TransFuser &58.37 &86.00 &0.67 \\
\hline
Ours &\bfseries69.17 &93.24 &\bfseries0.73 \\
\hline
\end{tabular}
\label{tab:2}
\end{center}
\end{table}

\section{Methodology}
The proposed method shown in Fig. \ref{fig1} comprises three main processes: 1) Extraction of spatial features from all modalities individually using CNNs; 2) Integration of sets of encodings to generate interpretable features; 3) Prediction of the forward waypoints utilizing the interpretable features. 

\subsection{Extraction of spatial features}
The camera images are inputted into the backbone network such as RegNetY-32 \cite{b8}. This process generates a feature map $F_{\text{camera}} \in \mathbb{R}^{C \times h_{\text{camera}} \times w_{\text{camera}}}$, where $C$ represents the number of channels in the feature map; $h_{\text{camera}} \times w_{\text{camera}}$ denotes the dimensions of the image-view features. For the LiDAR point clouds, previous works \cite{b15} are considered to encode the LiDAR point cloud data into a 3-bin histogram over a 3D BEV grid. Following the 3D backbone \cite{b8}, feature map $F_{\text{lidar}} \in \mathbb{R}^{C \times h_{\text{lidar}} \times w_{\text{lidar}}}$ is obtained.

\subsection{Integration of sets of encodings to generate interpretable features}
For encoding, the feature map $F$ from each sensor is processed using a $1 \times 1$ convolution to obtain a lower-channel feature map $f \in R^{c \times X \times Y}$, where $c$ is the desired number of output channels, and $X$ and $Y$ represent the spatial dimensions of the feature map. The $X$ and $Y$ are collapsed into one dimension, resulting in $c \times XY$ tokens. A fixed 2D sinusoidal positional encoding $e \in R^{c \times XY}$ is added to each token to preserve positional information. Additionally, a learnable sensor encoding $s \in R^{c \times N}$ dimensions is included to differentiate tokens from $N$ different sensors as follows

\begin{equation}
v_n^{(x,y)}=z_n^{(x,y)} + s_n + e^{(x,y)}\label{eq1}
\end{equation}
where $v_n$ represents encoded tokens, $z_n$ represents the tokens extracted from the $n$-th sensor, $x$ and $y$ denote the coordinates of each token obtained by each sensor. The encoded tokens from all sensors are concatenated and passed through a Transformer decoder. This enables the proposed framework to capture token relationships and interactions.

The decoder in the proposed framework takes a standard Transformer architecture\cite{b7}. Each decoder layer utilizes the queries to gather spatial information from the multi-modal features through the attention mechanism. The resulting outputs are reshaped into two feature maps with dimensions ${C\times h_{\text{camera}}\times w_{\text{camera}}}$ and ${C\times h_{\text{lidar}}\times w_{\text{lidar}}}$. These feature maps are combined with the existing feature map in each modality branch using element-wise summation. The compact representation of the environment is encoded in a 512-dimensional fused vector, capturing the global context of the 3D scene. 

\subsection{Forward waypoints prediction}
The Transformer decoder is accompanied by a prediction module, utilizing multiple Gated Recurrent Unit (GRU) \cite{b16} to forecast the waypoints. The 512-dimensional fused vector is passed through a multi-layer perceptron (MLP) to reduce its dimensionality to 64. The resulting vector is fed into GRU networks to predict the waypoints.

Similar to the training process in \cite{b17}, the network is trained using an L1 loss which measures the discrepancy between predicted and ground truth waypoints.
The loss function is defined as follows
\begin{equation}
L = \sum_{t=1}^{T} \|w_t - w^{gt}_t\|_1\label{eq2}
\end{equation}
where $w^{gt}_t$ represents the ground truth waypoint for the time step $t$.

\section{Experiments}
\subsection{Implementation details}
Training Dataset: The CARLA \cite{b18} simulator is used to train and test their autonomous driving model, with a dataset of images and point clouds collected from junctions and curved highways.

Benchmark:
Proposed method is evaluated with Longest6 benchmark \cite{b13} and Town05 Long benchmark \cite{b15}. The Longest6 benchmark comprises challenging driving conditions with high dynamic agent density, combining six weather and six daylight conditions. Town05 Long encompasses diverse road types, such as multi-lane roads, single-lane roads, bridges, highways, and exits. The primary focus of these benchmarks is to effectively manage dynamic agents and navigate through challenging adversarial events.

Metrics:
The proposed method is evaluated using three metrics: route completion (RC), infraction score (IS), and driving score (DS). RC measures the percentage of the route completed, IS decreases when infractions occur, and DS is a comprehensive metric that considers both progress and safety. We also show the additional metrics, e.g., Ped: Collisions with pedestrians, Veh: Collisions with vehicles, Stat: Collisions with static layout, Red: Red light violation, OR: Off-road driving, Dev: Route deviation, TO: Timeout, Block: Vehicle Blocked, for the infractions per kilometer metrics.

\subsection{Results with benchmarks}
Table \ref{tab:1} and Table \ref{tab:2} show the benchmark results of the proposed method and state-of-the-art approaches. The TransFuser \cite{b13} is a modality integration technique, excluding local and global embedding. Notably, we use their publicly available code to retrain the baseline model. The Latent TransFuser \cite{b13} adopts a similar architecture to the TransFuser but substitutes BEV LiDAR with fixed positional encoding image. Late Fusion \cite{b13} independently extracts image and point cloud features, then fuses them via element-wise summation. Geometric Fusion \cite{b13}, influenced by \cite{b22}, combines LiDAR and camera data through a multi-scale feature fusion using projection.

As seen in Table \ref{tab:1}, proposed method outperforms other models with the Longest6 benchmark in two major metrics, with the highest DS of 45.64 and the highest IS of 0.65. With Town05 Long benchmark, the results presented in Table \ref{tab:2} indicate the superior performance of the proposed method with top IS of 0.73. Other methods that prioritize reaching the goal at all cost may make more mistakes and violate traffic rules, resulting in lower IS scores. Our method takes into account the surrounding environment and makes more correct decisions, which enhances the safety of the driving process.

\section{Conclusion}
This work introduces a novel multi-modal fusion Transformer that effectively captures the overall 3D scene context, integrating local and global contexts. Compared with the TransFuser that can be taken as the state-of-the-art technique, the proposed method achieves 8\% and 19\% improvements in driving scores for the Longest6 and Town05 Long benchmarks, respectively.
\section*{Acknowledgment}
This work was supported by the Institute for Information communications Technology Promotion (IITP) grant funded by the Korean government (MSIT) (No.2020-0-00440, Development of Artificial Intelligence Technology that continuously improves itself as the situation changes in the real world).


\begin{thebibliography}{00}
\bibitem{b1}
A. Prakash, A. Behl, E. Ohn-Bar, K. Chitta, and A. Geiger, "Exploring data aggregation in policy learning for vision-based urban autonomous driving," in \textit{Proceedings of the IEEE/CVF Conference on Computer Vision and Pattern Recognition}, pp. 11763--11773, 2020.

\bibitem{b2}
A. Rudenko, L. Palmieri, M. Herman, K. M. Kitani, D. M. Gavrila, and K. O. Arras, "Human motion trajectory prediction: A survey," \textit{The International Journal of Robotics Research}, vol. 39, no. 8, pp. 895--935, 2020, Sage Publications Sage UK: London, England.

\bibitem{lai2022learning}
Q.-V. Lai-Dang, S. H. Nengroo, and H. Jin, "Learning Dense Features for Point Cloud Registration Using a Graph Attention Network," \textit{Applied Sciences}, vol. 12, no. 14, pp. 7023, 2022, MDPI.

\bibitem{b3}
S. Y. Alaba and J. E. Ball, "Deep Learning-based Image 3D Object Detection for Autonomous Driving," \textit{IEEE Sensors Journal}, 2023, IEEE.

\bibitem{rajendran2022relmobnet}
P. K. Rajendran, S. Mishra, L. F. Vecchietti, and D. Har, "RelMobNet: End-to-end relative camera pose estimation using a robust two-stage training," in \textit{European Conference on Computer Vision}, pp. 238--252, 2022, Springer.

\bibitem{rajendran2022lightweight}
P. K. Rajendran, Q. V. Lai-Dang, L. F. Vecchietti, and D. Har, "A Lightweight Domain Adaptive Absolute Pose Regressor Using Barlow Twins Objective," \textit{arXiv preprint arXiv:2211.10963}, 2022.

\bibitem{Park1}
I. Park, D. Kim, and D. Har, "MAC achieving low latency and energy efficiency in hierarchical M2M networks with clustered nodes," \textit{IEEE Sensors Journal}, pp. 1657-1661, 2015.

\bibitem{Park2}
J. Park, E. Hong, and D. Har, "Low complexity data decoding for SLM-based OFDM systems without side information," \textit{IEEE Communications Letters}, pp.611-613, 2011.

\bibitem{b4}
Z. Huang, C. Lv, Y. Xing, and J. Wu, "Multi-modal sensor fusion-based deep neural network for end-to-end autonomous driving with scene understanding," \textit{IEEE Sensors Journal}, vol. 21, no. 10, pp. 11781--11790, 2020, IEEE.

\bibitem{b17}
Dian Chen, Brady Zhou, Vladlen Koltun, and Philipp Kr{\"a}henb{\"u}hl.
"Learning by cheating."
\textit{Conference on Robot Learning}, pp. 66--75, 2020.
PMLR.

\bibitem{b5}
K. Chitta, A. Prakash, and A. Geiger, "Neat: Neural attention fields for end-to-end autonomous driving," in \textit{Proceedings of the IEEE/CVF International Conference on Computer Vision}, pp. 15793--15803, 2021.

\bibitem{b6}
K. Ishihara, A. Kanervisto, J. Miura, and V. Hautamaki, "Multi-task learning with attention for end-to-end autonomous driving," in \textit{Proceedings of the IEEE/CVF Conference on Computer Vision and Pattern Recognition}, pp. 2902--2911, 2021.

\bibitem{kim2020machine}
T. Kim, L. F. Vecchietti, K. Choi, S. Lee, and D. Har, "Machine learning for advanced wireless sensor networks: A review," \textit{IEEE Sensors Journal}, vol. 21, no. 11, pp. 12379--12397, 2020.

\bibitem{b7}
A. Vaswani, N. Shazeer, N. Parmar, J. Uszkoreit, L. Jones, A. N. Gomez, {\L}. Kaiser, and I. Polosukhin, "Attention is all you need," \textit{Advances in neural information processing systems}, vol. 30, 2017.

\bibitem{b8}
I. Radosavovic, R. P. Kosaraju, R. Girshick, K. He, and P. Doll{\'a}r, "Designing network design spaces," in \textit{Proceedings of the IEEE/CVF conference on computer vision and pattern recognition}, pp. 10428--10436, 2020.

\bibitem{b9}
S. Pang, D. Morris, and H. Radha, "CLOCs: Camera-LiDAR object candidates fusion for 3D object detection," in \textit{2020 IEEE/RSJ International Conference on Intelligent Robots and Systems (IROS)}, pp. 10386--10393, 2020, IEEE.

\bibitem{b10}
S. Vora, A. H. Lang, B. Helou, and O. Beijbom, "Pointpainting: Sequential fusion for 3D object detection," in \textit{Proceedings of the IEEE/CVF conference on computer vision and pattern recognition}, pp. 4604--4612, 2020.

\bibitem{b11}
X. Chen, H. Ma, J. Wan, B. Li, and T. Xia, "Multi-view 3D object detection network for autonomous driving," in \textit{Proceedings of the IEEE Conference on Computer Vision and Pattern Recognition}, pp. 1907--1915, 2017.

\bibitem{b12}
Z. Liu, H. Tang, A. Amini, X. Yang, H. Mao, D. Rus, and S. Han, "BEVFusion: Multi-Task Multi-Sensor Fusion with Unified Bird's-Eye View Representation," \textit{arXiv preprint arXiv:2205.13542}, 2022.

\bibitem{b13}
K. Chitta, A. Prakash, B. Jaeger, Z. Yu, K. Renz, and A. Geiger, "Transfuser: Imitation with transformer-based sensor fusion for autonomous driving," \textit{IEEE Transactions on Pattern Analysis and Machine Intelligence}, 2022.

\bibitem{b15}
A. Prakash, K. Chitta, and A. Geiger, "Multi-modal fusion transformer for end-to-end autonomous driving," \textit{Proceedings of the IEEE/CVF Conference on Computer Vision and Pattern Recognition}, pages 7077--7087, 2021.

\bibitem{b16}
K. Cho, B. V. Merrienboer, C. Gulcehre, D. Bahdanau, F. Bougares, H. Schwenk, and Y. Bengio, "Learning phrase representations using RNN encoder-decoder for statistical machine translation," \textit{arXiv preprint arXiv:1406.1078}, 2014.


\bibitem{b18}
A. Dosovitskiy, G. Ros, F. Codevilla, A. Lopez, and V. Koltun, "CARLA: An open urban driving simulator," \textit{Conference on Robot Learning}, pp. 1--16, 2017, PMLR.

\bibitem{b21}
S. Vora, A. H. Lang, B. Helou, and O. Beijbom, "Pointpainting: Sequential fusion for 3d object detection," \textit{Proceedings of the IEEE/CVF Conference on Computer Vision and Pattern Recognition}, pp. 4604--4612, 2020.

\bibitem{b22}
M. Liang, B. Yang, S. Wang, and R. Urtasun, "Deep continuous fusion for multi-sensor 3D object detection," \textit{Proceedings of the European Conference on Computer Vision (ECCV)}, pp. 641--656, 2018.

\end{thebibliography}
\end{document}